# Sharpening the Spear: Adaptive Expert-Guided Adversarial Attack Against DRL-based Autonomous Driving Policies

Junchao Fan, Xuyang Lei, Xiaolin Chang, *Senior Member*, *IEEE*

*Abstract*—Deep reinforcement learning (DRL) has emerged as a promising paradigm for autonomous driving. However, despite their advanced capabilities, DRL-based policies remain highly vulnerable to adversarial attacks, posing serious safety risks in real-world deployments. Investigating such attacks is crucial for revealing policy vulnerabilities and guiding the development of more robust autonomous systems. While prior attack methods have made notable progress, they still face several challenges: 1) they often rely on high-frequency attacks, yet critical attack opportunities are typically context-dependent and temporally sparse, resulting in inefficient attack patterns; 2) restricting attack frequency can improve efficiency but often results in unstable training due to the adversary's limited exploration. To address these challenges, we propose an adaptive expert-guided adversarial attack method that enhances both the stability and efficiency of attack policy training. Our method first derives an expert policy from successful attack demonstrations using imitation learning, strengthened by an ensemble Mixture-of-Experts architecture for robust generalization across scenarios. This expert policy then guides a DRL-based adversary through a KL-divergence regularization term. Due to the diversity of scenarios, expert policies may be imperfect. To address this, we further introduce a performance-aware annealing strategy that gradually reduces reliance on the expert as the adversary improves. Extensive experiments demonstrate that our method achieves outperforms existing approaches in terms of collision rate, attack efficiency, and training stability, especially in cases where the expert policy is sub-optimal.

## I. INTRODUCTION

IN recent years, deep reinforcement learning (DRL) has emerged as a powerful paradigm for tackling the complex decision-making problems encountered in autonomous driving [1]-[3]. Its ability to learn from high-dimensional, sequential interactions makes it particularly well-suited for handling the uncertainty, variability, and long-term dependencies characteristic of real-world driving environments. This potential has sparked widespread interest in actively integrating DRL into the development process to achieve the next generation of autonomous systems [4].

Despite these promising developments, the real-world deployment of DRL remains hindered by critical safety and robustness concerns, particularly due to vulnerabilities to adversarial attacks [5][6]. Studies have shown that neural network-based decision policies can suffer substantial performance degradation under input perturbations or environmental disturbances, potentially leading to safety incidents [7]. Ensuring the reliability of DRL under such perturbations is therefore essential for its safe adoption in safety-critical domains like autonomous driving.

In this context, systematically identifying and analyzing the vulnerabilities of DRL-based driving policies becomes a crucial step toward building safer and more robust autonomous systems. Recent efforts have begun to explore adversarial attacks on DRL-based autonomous driving policies [8]-[10]. However, most existing methods apply high-frequency perturbations throughout the driving process, overlooking the fact that truly critical attack opportunities are often sparse and context-dependent. Such indiscriminate attack strategies not only incur excessive computational and attack costs, making it difficult to iterate or generalize defense policies effectively. Moreover, adversarial perturbations throughout the entire process can easily cause defense strategies to become overly conservative.

To address this, some recent works [11][12] have proposed DRL-based adversaries that aim to identify and exploit optimal attack timing. For instance, Fan *et al.* [13] introduced a DRL-based adversary capable of autonomously decide the timing and content of attacks under strict attack frequency constraints, aiming to inducing collisions in autonomous driving scenarios. While these approaches show promise in improving attack efficiency, the imposed constraints can also introduce training instability. Specifically, by limiting the number or frequency of attacks, such constraints inadvertently restrict the adversary's exploration ability, making it harder to learn effective policies, particularly in complex and sparse-reward autonomous driving environments.

In this paper, we propose a novel expert-guided DRL-based adversarial attack method designed to improve both the training stability and attack effectiveness of the adversary under strict attack frequency constraints. Our approach consists of two main stages. In the first stage, we leverage imitation learning to derive an expert attack policy from a collection of successful attack demonstrations. We further adopt a Mixture-of-Experts (MoE) architecture into the imitation learning framework to enhance generalization ability of the expert across diverse driving scenarios. In the second stage, the expert policy is used to guide the learning of a DRL-based adversary. Specifically, we introduce a regularization term based on the Kullback-Leibler (KL) divergence between the expert and adversary policies. This term is incorporated into the adversary's optimization objective to constrain its behavior within a desirable action space. Unlike the inefficient trial-and-error exploration of vanilla DRL methods, this approach improves both training efficiency and stability. However, the expert policy may be sub-optimal in diverse or unseen scenarios [1]. To address this, we further incorporate a performance-aware annealing mechanism that progressively reduces expert reliance as the adversary improves. This enables the adversary to flexibly balance expert guidance and autonomous exploration, facilitating the learning of more effective attack policies. The main contributions of this paper are listed as follows:

1) **Expert-Guided Adversarial Attack Framework**: We propose a novel expert-guided adversarial attack method for autonomous driving, which enhances existing low-frequency attack frameworks by incorporating expert knowledge to improve both training stability and attack performance. To the best of our knowledge, we are the first to integrate expert knowledge into adversarial attack training for DRL-based autonomous driving policies.
2) **Mixture-of-Experts-Based Expert Policy Derivation**: To improve the quality and generalizability of expert guidance, we introduce a Mixture-of-Experts (MoE) architecture to derive the expert policy from historical successful attack demonstrations. This design enables the expert to better capture diverse attack policies across various scenarios.
3) **Performance-aware Annealing for Adaptive Expert Reliance**: To address the limitations of imperfect expert policies, we introduce a performance-aware annealing mechanism that dynamically adjusts the reliance on the expert based on the adversary's performance during training. This mechanism enables the adversary to flexible balance expert guidance with autonomous exploration, smoothly adapting to varying levels of expert quality to learn more effective policies.

Experimental results demonstrate that our method consistently outperforms existing adversarial attack baselines across multiple autonomous driving scenarios. Specifically, our approach achieves significantly higher collision rates and attack efficiency under strict budget constraints, while also exhibiting faster convergence and more stable training dynamics.

The remaining paper is structured as follows. Section II introduces the related work. Section III outlines the preliminaries on DRL and formulates the attack problem. In Section IV, we detail our proposed attack method. The experimental results are discussed and analyzed in Section V, followed by the conclusion in Section VI.

## II. RELATED WORK

This section first reviews the recent research on DRL-based autonomous driving policies in Section II.A, followed by a review of adversarial attack methods targeting these policies in Section II.B. Section II.C summarizes the gaps between our work and existing attack methods.

*A. DRL-based Autonomous Driving*

Driven by rapid advancements in machine intelligence and data-driven approaches, Deep Reinforcement Learning (DRL) has emerged as a forefront solution for autonomous driving [1]. For instance, Lu *et al.* [16] proposed a data-rule fusion-driven DRL algorithm to overcome the challenges of manually designing reward functions in complex environments. Huang *et al.* [17] integrated human prior knowledge with DRL to enhance algorithmic performance, and validated their method in unprotected left-turn and roundabout tasks. Li *et al.* [18] proposed a hierarchical skill-based offline reinforcement learning method to solve the long-horizon vehicle planning task. Dang *et al.* [19] developed an event-triggered model predictive control framework based on DRL to address the path following problem in autonomous driving.

Although significant progress has been achieved, the absence of safety guarantees continues to pose substantial constraints on the real-world deployment of above methods. The exploration-driven and reward-based nature of DRL makes safety verification crucial, as its opaque policies may cause catastrophic failures in safety-critical driving scenarios. Some studies have attempted to improve the safety of DRL-based autonomous driving policies by introducing external risk assessment methods, such as safety checkers [20], safety constraints [21] and human feedback [22]-[24]. For instance, He *et al.* [25] proposed a fear model to estimate the dangers of current state, thereby guiding the DRL agent to adopt safer driving policies. Li *et al.* [26] proposed a novel paradigm of expert intervention, enabling human experts to take control in hazardous situations and demonstrate safe driving behaviors to help the agent learn safer driving policies. Chen *et al.* [27] proposed an external safety layer that evaluates the risk of actions generated by a DRL-based motion planning controller, thereby enhancing safety. While above methods significantly improved the safety in ideal, noise-free environments, they may fail to ensure the robustness and safety in the presence of adversarial perturbations.

*B. Adversarial Attacks for DRL-based Autonomous Driving*

Existing studies have demonstrated that well-trained DRL policies remain highly vulnerable to adversarial perturbations [29]. These carefully crafted perturbations can be injected into the autonomous driving agent's observations to induce hazardous behaviors (e.g., sudden emergency braking) [30]-[35]. For instance, Wang *et al.* [36] investigated an adversary which can introduce adversarial perturbations into the action space of the autonomous driving agent.

While above attack methods can achieve some success, they that adversarial perturbations are applied continuously throughout the driving process. In other words, they overlook the fact that critical attack opportunities are inherently sparse during the driving process. For instance, in tasks like intersection navigation, the optimal timing for adversarial attacks often coincides with the crossing phase, when traffic is most dynamic and unpredictable. To address this, recent studies have begun exploring more temporally targeted attack policies to improve efficiency [37]-[39]. Specifically, these attacks limit the number of attack times with the aim of encouraging the adversary to learn when to launch attacks. Due to the need for self-learning, these adversaries are typically based on DRL. Although such methods demonstrated the ability to achieve comparable collision rates while significantly reducing attack frequency compared to high-frequency attacks, they still have limitations. Specifically, limiting the attack frequency restricts the adversary's exploration capabilities. Coupled with DRL's inherently low sample efficiency, this often leads to failure in complex scenarios. Different from making the adversary learns attack policies from scratch, we take a step further by reusing successful past experiences. Specifically, we leverage expert knowledge distilled from prior tasks to guide adversary training in new environments, resulting in more efficient and stable learning.

*C. RL with Expert Prior Knowledge*

Although DRL is regarded as a promising paradigm for tackling complex decision-making problems, its effectiveness remains constrained by inherent limitations such as low sample efficiency. To address this challenge, many studies incorporate expert prior knowledge (e.g., rules, demonstrations, and interventions) into the DRL training framework for autonomous driving policies [40][41]. For instance, Li *et al.* [12] proposed a novel human-in-the-loop learning method to enable human experts to intervene in the behavior of autonomous driving agents. However, one limitation of the human-in-the-loop paradigm is its reliance on real-time human supervision. In contrast, using offline expert demonstrations to guide training is more practical in many scenarios. For instance, Huang *et al.* [42] employed behavior cloning to derive an imitative expert policy from human expert demonstrations, which then regularizes the RL agent's behavior.

*D. Research Gaps*

In contrast to existing work, our research focuses on addressing the following gaps:
1) **Overly Frequent Adversarial Attacks**: Existing high-frequency attack methods typically apply perturbations continuously over time. However, effective attack opportunities are temporally sparse and context-dependent. These methods not only waste computational resources but also limit the ability to expose key vulnerabilities, ultimately hindering progress toward safer and more robust autonomous driving policies.
2) **Exploration Inefficiency of DRL-based method**: Existing low-frequency methods typically rely on DRL to explore adversarial policies from scratch. However, such trial-and-error exploration is highly inefficient, especially under strict attack budgets and sparse rewards.
3) **Ignoring Expert Suboptimality**: Existing expert-guided DRL methods often assume the expert is near-optimal and overlook that, in our setting, the expert is inherently imperfect due to the diversity of environments and attack conditions. This suboptimality can mislead adversary training and limit its overall effectiveness if left unaddressed.

Based on these considerations, we propose an efficient adaptive expert-guided adversarial attack method.

III. PRELIMINARY

This section first introduces the fundamental principles of reinforcement learning in Section III.A, followed by the formulation of the adversarial attack problem in Section III.B.

*A. Deep Reinforcement Learning*

Before applying DRL algorithms to learn a policy, the decision-making problem needs to be formulated as a Markov Decision Process (MDP). An MDP can be defined by a tuple $(\mathcal{S}, \mathcal{A}, r, \mathcal{P}, \gamma)$. Here, $\mathcal{S}$ and $\mathcal{A}$ denote the state space and action space, respectively, representing the sets of all possible system states and control actions. In autonomous driving systems, the state may include information about the ego vehicle and surrounding vehicles. The action space may consist of control commands such as steering and braking. The reward function $r: \mathcal{S} \times \mathcal{A} \to \mathbb{R}$ is designed to guide the agent learn the desired policy and $r(s,a)$ is the immediate reward received after taking action $a$ in state $s$. $\mathcal{P}: \mathcal{S} \times \mathcal{A} \times \mathcal{S} \to [0,1]$ is the transition probability distribution of the system, where $p(s'|s,a)$ gives the probability of transitioning to from state $s$ to state $s'$ when taking action $a$. $\gamma \in (0,1)$ is the discount factor which control the trade-off between immediate rewards and future returns. The goal of DRL is to find an optimal policy $\pi^*$ that maximizes the expected cumulative discounted reward

$$J(\pi) = \mathbb{E}_{a_t \sim \pi(s_t)} \left( \sum_{t=1}^{T} \gamma^t r(s_t, a_t) \right). \tag{1}$$

*B. Problem Formulation*

Existing work [A] proposed a novel adversarial attack that significantly reduces the attack frequency. This low-frequency attack policy reduces the computational overhead caused by frequent perturbations, thereby indirectly enhancing the efficiency of training defense models. Building upon this foundational method, our method further enhances training efficiency by incorporating expert prior knowledge.

The goal of the low-frequency attack method is to maximizes the expected cumulative adversarial reward under constraint of limited attack budget. We thus formulate the attack as the following optimization problem:

$$\max_{(x, a_{adv}'')} \mathbb{E}_{a_t'' \sim \pi(s_t')} \left( \sum_{t=1}^{T} \gamma^t \overline{r}(s_t', a_t'') \right), \tag{2}$$

s.t.
$$\delta_t = PG(\pi, s_t, a_t'), t \in T, \tag{2a}$$
$$s_t' = s_t + x_t \delta_t, t \in T, \tag{2b}$$
$$\|\delta_t\|_\infty \leq \epsilon, t \in T, \tag{2c}$$
$$\sum_{t \in T} x_t \leq \Gamma, \Gamma \ll T. \tag{2d}$$

Eq. (2) is the objective function that maximizes the adversary's cumulative discounted reward over the entire episode, where $T$

is the maximum length of each episode. $\bar{r}$ is the reward function of the adversary, which detailed in Section V.B. The use of cumulative discounted rewards implicitly encourages the adversary to achieve its goal in as few time steps as possible. The optimization variables $(x, a_{adv})$ consists of two components. $x = \{x_t, t \in T\}$ is a set of binary variables that determine whether an adversarial attack is launched at each time step. $a_{adv} = \{a'_t, t \in T\}$ denotes the actions that the adversary expects the victim agent to take, which are used to guide perturbation generation. When $x_t = 1$, the adversary invokes a perturbation generation function ($PG$) to generate perturbation $\delta_t$. As shown in (2a), the inputs of $PG$ includes the agent's policy $\pi$, the current state $s_t$, and $a'_t$. This perturbation is then injected into the environment, as demonstrated by (2b). The agent then selects an action $a''_t$ according to $s'_t$. To ensure the perturbations are reasonable and consistent with the dynamics of real-world environments, we impose an $\ell_\infty - norm$ constraint on $\delta_t$, as shown in (2c). Eq. (2d) ensures that the total number of adversarial perturbations across the episode remains below a predefined limit $\Gamma$, which denoted as the attack budget. Solving this optimization problem is particularly challenging due to its non-linear dynamics and mixed-integer formulation, with complexity increasing rapidly as the time horizon $T$ grows.

IV. METHODOLOGY

This section first outlines the overall framework of our attack method in Section IV.A, followed by a detailed description of the algorithm implementation and the trajectory clipping technique in Sections IV.B and Section IV.C, respectively.

*A. Framework*

The overall framework of our method is illustrated in Fig. 1 and consists of three stages: (a) demonstration data collection, (b) expert policy training, and (c) expert-guided DRL training, which are detailed in Sections III.B-III.D. First, in the demonstration data collection stage, we gather existing adversarial trajectories and encode them into state–action pairs $(s_t^{adv}, a_t^{adv})$. The resulting set of pairs serves as the training dataset $\mathcal{D}_E : \{(s_t^{adv}, a_t^{adv})\}_N$ for generating the expert policy in the second stage. Next, in the export policy training stage, we employ behavior cloning (BC) to train an adversarial expert policy $\pi_\theta^e : \mathcal{S} \to \mathcal{A}$. Finally, during the expert-guided DRL training stage, we use the Kullback–Leibler (KL) divergence between the expert and adversary policies as a regularization term to guide the training process of the adversary. An adaptive threshold dynamically adjusts its weight to balance expert guidance and policy exploration over time. This mechanism encourages early reliance on the expert for stable learning, gradually reducing it to foster independent policy refinement and enhanced attack performance.

*B. Demonstration Data Collection*

Since adversarial trained models can systematically generate challenging attack patterns, we employ them as experts. Specifically, we employ multiple attack models trained using the attack algorithm from [13], aggregating their attack trajectories to form a high-quality expert demonstration dataset $\mathcal{D}_E$. Each sample $(s_t^{adv}, a_t^{adv}) \in \mathcal{D}_E$ is a state-action pair, where the state $s_t^{adv}$ includes the observation $s_t$ of the ego vehicle, its original(pre-attack) action $a_t$, and the remaining attack steps $n_t$. The action $a_t^{adv} = (x_t, a'_t)$ is a two-dimensional vector, as detailed in Section III.B.

As we intend to employ behavior cloning to derive an expert policy in the subsequent stage, and given that behavior cloning demands high-quality demonstration samples, it is essential to ensure the reliability of the data. To this end, we carefully preprocess the demonstration data before training. First, we filter out failed episodes and retain only the state–action pairs from successful attack demonstrations to avoid misleading the learning process. Second, we randomly undersample non-attack samples to mitigate the heavy imbalance between attack and non-attack samples. Finally, to improve generalization and prevent overfitting to a single scenario, we incorporate attack trajectories from multiple scenarios so the expert can learn more robust behaviors [43].

*C. Expert Policy Training Stage*

Given the sparsity and contextual complexity of adversarial attacks, expert behavior is difficult to define manually or through rule-based heuristics. To address this, we employ behavior cloning to learn an expert policy from the high-quality trajectories constructed in Section IV.B. This data-driven approach enables the policy to capture implicit attack strategies, leading to improved robustness and generalization. The expert policy $\pi_\theta^e : \mathcal{S} \to \mathcal{A}$ can be obtained by maximizing the likelihood of expert actions conditioned on observed states, as follows:

$$\pi_\theta^e \coloneqq \arg\max_\pi \mathbb{E}_{(s_t^{adv}, a_t^{adv}) \sim \mathcal{D}^E} \left[ \log \pi_\theta(a_t^{adv} | s_t^{adv}) \right] \quad (3)$$

To capture the uncertainty inherent in expert behavior, we model the expert policy $\pi_\theta^e(s_t^{adv})$ as a state-dependent Gaussian distribution $\mathcal{N}\left(\hat{\mu}_\theta(s_t^{adv}), \hat{\sigma}_\theta^2(s_t^{adv})\right)$ rather than a deterministic mapping. $\hat{\mu}_\theta$ and $\hat{\sigma}_\theta^2$ are the predictive mean and variance, respectively. Under this probabilistic formulation, maximizing the log-likelihood is equivalent to minimizing the negative log-likelihood loss [46]:

$$\mathcal{L}(\theta) = \mathbb{E}_{\mathcal{D}^E} \left[ \frac{\log \hat{\sigma}_\theta^2(s_t^{adv})}{2} + \frac{(a_t^{adv} - \hat{\mu}_\theta(s_t^{adv}))^2}{2\hat{\sigma}_\theta^2(s_t^{adv})} + c \right] \quad (4)$$

where $\theta$ is the parameters of the expert policy and $c$ is a constant.

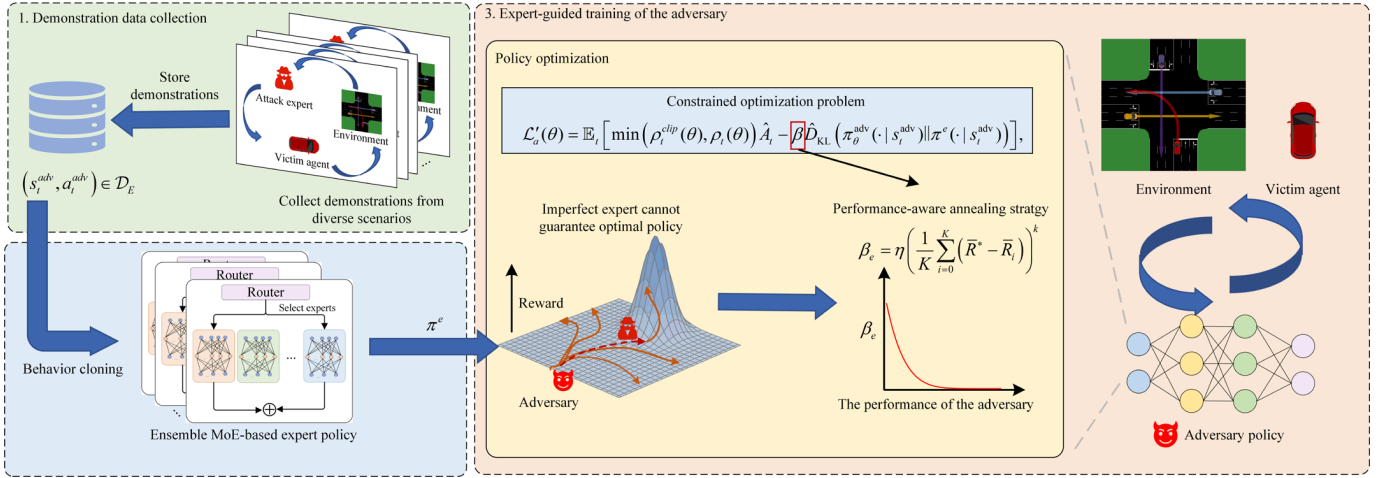

**Fig. 1.** The framework of our attack method

Since the demonstration data are collected from a diverse range of environments, we employ a Mixture-of-Experts (MoE) [44][45] architecture to better capture this variability. By allowing different experts to specialize in distinct contexts, MoE enables the policy to model complex behaviors more effectively, leading to stronger generalization and greater robustness across varied scenarios. Specifically, a MoE network consists of multiple expert networks and a router network. Each expert processes the same input and produces its own output, while a router network assigns normalized weights to each expert based on the input. Specifically, for a given input state $s_t^{\text{adv}}$, each expert network $\pi^{e_i}$ outputs action distribution $\mathcal{N}\left(\mu_i\left(s_t^{\text{adv}}\right), \sigma_i\left(s_t^{\text{adv}}\right)\right)$. The router network outputs a weight vector for the experts:

$$w\left(s_t^{\text{adv}}\right) = \left[w_1\left(s_t^{\text{adv}}\right), w_2\left(s_t^{\text{adv}}\right), ..., w_N\left(s_t^{\text{adv}}\right)\right] \quad (5)$$

where $N$ denotes the number of experts. The final ouput of the MoE network is the weighted sum of the experts' outputs:

$$\mu = \sum_{i=1}^{M} w_i \mu_i, \quad \sigma = \sum_{i=1}^{M} w_i \sigma_i \quad (6)$$

Due to the limited scale of the demonstration data, the MoE-based expert may still become unreliable when facing unseen states. To improve its generalization and robustness, we introduce an ensemble-based MoE architecture that integrates multiple MoE networks. Specifically, we train $M$ MoE networks with random initializations and aggregate their outputs into a Gaussian mixture distribution:

$$\mu_*(s_t^{\text{adv}}) = \frac{1}{M} \sum_{i=1}^{M} \hat{\mu}_{\theta_i}(s_t^{\text{adv}}) \quad (7)$$

$$\sigma_*^2(s_t^{\text{adv}}) = \frac{1}{M} \sum_{i=1}^{M} \left(\hat{\sigma}_{\theta_i}^2\left(s_t^{\text{adv}}\right) + \hat{\mu}_{\theta_i}^2\left(s_t^{\text{adv}}\right)\right) - \mu_*^2(s_t^{\text{adv}}) \quad (8)$$

When $\sigma_*^2(s_t^{\text{adv}})$ is small, it indicates high ensemble confidence, encouraging the adversarial agent to align closely with the expert distribution. In contrast, a large $\sigma_*^2(s_t^{\text{adv}})$ promotes exploration within the constrained policy space, potentially yielding improved attack policies [46].

*D. Expert-Guided Training of the Adversary*

In this section, we leverage the trained ensemble expert policy to guide the learning of the adversary in new cases. To achieve stable and efficient policy updates, we adopt Proximal Policy Optimization (PPO), a state-of-the-art policy-gradient method known for balancing sample efficiency with training stability [47][48]. We further modify the update mechanism of PPO to incorporate expert knowledge, enabling expert-guided learning during adversary training. Our algorithm consists of two networks, an actor network $\pi_\theta^{adv}$ and a critic network $V_\phi^{adv}$, where $\theta$ and $\phi$ are the parameters corresponding to these two networks.

Given an observation $s_t^{\text{adv}}$, $\pi_\theta^{adv}$ will output a Gaussian distribution over actions. The objective of $\pi_\theta^{adv}$ is to maximize the expect cumulative reward by improving the quality of the actions selected. Specifically, $\pi_\theta^{adv}$ is optimized using the following loss function:

$$\mathcal{L}_a(\theta) = \mathbb{E}_t\left[\min\left(\rho_t^{clip}(\theta), \rho_t(\theta)\right) \hat{A}_t\right], \quad (9)$$

where

$$\rho_t^{clip}(\theta) = clip\left(\rho_t(\theta), 1-\varepsilon, 1+\varepsilon\right), \quad (10)$$

$$\rho_t(\theta) = \frac{\pi_\theta^{adv}\left((p_t, a_t') \mid s_t^{adv}\right)}{\pi_\theta^{adv,old}\left((p_t, a_t') \mid s_t^{adv}\right)}, \quad (11)$$

where $\rho_t(\theta)$ represents the probability ratio between the new and old policies. $\varepsilon$ is the clipping parameter which serves to constrain the policy update step and prevent excessively large deviations from the old policy. $\hat{A}_t$ is the estimated advantage and can be defined as

$$\hat{A}_t = \sum_{i=t}^{T-1}(\gamma\lambda)^{i-t}\left(\overline{r}_t + \gamma V_\phi^{adv}(s_{t+1}^{adv}) - \gamma V_\phi^{adv}(s_t^{adv})\right), \quad (12)$$

where $\lambda$ is a tuning parameter.

The critic network $V_\phi^{adv}$ estimates the state value function by minimizing the discrepancy between its predictions and empirical returns. This can be achieved by minimizing the mean squared error (MSE) between the predicted value and the target

return, as follows:

$$\mathcal{L}_c(\phi) = \mathbb{E}_t \left[ \left( V_\phi^{adv}(s_t^{adv}) - \left( \bar{r}_t + \gamma V_\phi^{adv}(s_{t+1}^{adv}) \right) \right)^2 \right]. \quad (13)$$

Then we can obtain the total loss function as

$$\mathcal{L}_{total}(\theta, \phi) = \mathcal{L}_a(\theta) - c_1 \mathcal{L}_c(\phi) + c_2 \mathcal{H}\left[ \pi_\theta^{adv}(\cdot \mid s_t^{adv}) \right], \quad (14)$$

where $\mathcal{H}\left[ \pi_\theta^{adv}(\cdot \mid s_t^{adv}) \right]$ is an entropy regularization term to encourage exploration. $c_1$ and $c_2$ are coefficients. $\theta$ and $\phi$ can be simultaneously updated using gradient-based methods.

Building on the standard PPO update framework, we proceed to integrate expert knowledge into the training process of the adversary. Specifically, we incorporate the KL-divergence between the adversary's policy and the expert policy as a constraint in $\mathcal{L}_a(\theta)$, formulated as follows:

$$\mathcal{L}_a'(\theta) = \mathbb{E}_t \left[ \begin{array}{c} \min\left( \rho_t^{clip}(\theta), \rho_t(\theta) \right) \hat{A}_t - \\ \beta \hat{D}_{KL}\left( \pi_\theta^{adv}(\cdot \mid s_t^{adv}) \| \pi^e(\cdot \mid s_t^{adv}) \right) \end{array} \right], \quad (15)$$

where $\beta$ is the weighting coefficient of the KL-divergence. This divergence continuously guides the adversary's policy toward the expert policy, effectively distilling knowledge from the expert. As mentioned in [51] (refer to Lemma 2), when the adversarial policy underperforms compared to the expert policy, minimizing their KL divergence ensures a non-negative improvement in the adversary's performance.

**Lemma 1** (Policy Improvement Bound [51]). *When the agent's policy $\pi$ is suboptimal compared to the expert's policy $\pi^e$, for any new policy $\pi'$ that is also inferior to the expert policy must satisfies the following condition:*

$$J(\pi') - J(\pi) \geq \frac{\xi - \max_{s,a} |A^\pi(s,a)| \sqrt{2 D_{KL}(\pi' \| \pi^e)}}{1 - \gamma} \quad (16)$$

Nonetheless, the expert policy may be suboptimal in certain scenarios. This issue is more pronounced in our problem, where high-quality expert demonstrations are often limited, making it difficult to derive an expert policy that performs well across all scenarios. As the adversarial policy surpasses the expert during training, maintaining the constraint may no longer be beneficial and could potentially impede further improvement. Then we theoretically analyze how the constraint influence the policy update.

**Proposition 1.** *For a fixed coefficient $\beta$, the policy that maximizes the regularized objective is necessarily suboptimal, i.e., $J(\hat{\pi}_{reg}) < J(\pi^*)$.*

*Proof.* Let $\pi^*$ be the optimal policy for $J(\pi)$, and $\pi^e$ be the suboptimal expert policy where $\pi^e \neq \pi^*$. We prove by contradiction.
Assume that the maximizer of the regularized objective is the optimal policy $\hat{\pi}_{reg} = \pi^*$. Then the gradient of the regularized objective must be zero at $\hat{\pi}_{reg}$:

$$\nabla J(\hat{\pi}_{reg}) - \beta \nabla D_{KL}(\hat{\pi}_{reg} \| \pi^e) = 0 \quad (17)$$

Substituting our assumption $\hat{\pi}_{reg} = \pi^*$ into (15), we have:

$$\nabla J(\pi^*) - \beta \nabla D_{KL}(\pi^* \| \pi^e) = 0 \quad (18)$$

By definition, $\pi^*$ is the maximizer of $J(\pi)$, so we have:

$$\nabla J(\pi^*) = 0 \quad (19)$$

Substituting (17) into (16) yields:

$$\beta \nabla D_{KL}(\pi^* \| \pi^e) = 0 \quad (20)$$

Since $\beta \neq 0$, this requires $\nabla D_{KL}(\pi^* \| \pi^e) = 0$. The gradient of the KL divergence is zero if and only if the two distributions are identical. This would mean

$$\pi^* = \pi^e \quad (21)$$

This contradicts our initial premise that the expert policy $\pi^e$ is suboptimal and thus $\pi^e \neq \pi^*$. Hence, the initial assumption that $\hat{\pi}_{reg} = \pi^*$ must be false. It follows that $J(\hat{\pi}_{reg}) < J(\pi^*)$.
□

We then propose a performance-aware annealing strategy that enables the adversary to benefit from early expert guidance while avoiding long-term suboptimality. Specifically, we design an adaptive $\beta_e$ to dynamically balances expert guidance and policy exploration. The update rule is formulated as follows:

$$\beta_e = \eta \left( \frac{1}{K} \sum_{i=0}^{K} (\bar{R}^* - \bar{R}_i) \right)^k \quad (22)$$

where $\bar{R}^*$ denotes the adversary's optimal return per episode and $\bar{R}_i = \sum_{t=0}^{T} r_t$ denotes the adversary's actual total return obtained in the $i-th$ episode of the last iteration. $k$ serves as a annealing factor that controls the sensitivity of the variation, and $\eta$ is a scaling factor. Eq. (20) adapts $\beta_\tau$ according to the reward gap between the average episode return achieved in the last iteration and the optimal return. With continued training, the adversary's performance improves, as indicated by an increasing return $\bar{R}$. This leads to a corresponding decrease in the regularization coefficient $\beta_\tau$, gradually reducing the adversary's dependence on the expert and promoting autonomous exploration of more optimal policies.

The step-by-step implementation of our method is outlined in **Algorithm 1**. At each training iteration, the adversary collects trajectories by interacting with both the environment and the autonomous driving agent (Lines 1-16). Specifically, it generates an adversarial action $(x_t^{adv}, a_t^{adv})$ based on the current state $s_t$, the agent's initial action $a_t$, and the remaining attack budget $n_t$. If $x_t^{adv} > 0$ and the adversary still has remaining attack budget, an attack is launched, as illustrated in Lines 5–8. Conversely, if the conditions are not satisfied, the adversary remains inactive and allows the agent to proceed unperturbed (Lines 9-10). After the agent executes the action and interacts with the environment, the resulting transition is stored in the buffer (Lines 11-13). Here, *done* a termination signal returned by the environment to indicate whether the current episode has ended due to either a collision or task completion. When either the episode terminates (i.e., *done* is True) or the attack budget is exhausted, the environment is reset to start a new episode (Lines 14-15). Once the buffer is filled, we begin the training phase and perform multiple epochs of mini-batch gradient-based optimization (Lines 17-23), following the update rules defined in (10)–(13) and (20).

| Algorithm 1 Our attack method |
|---|
| **Input**: Environment $Env$, victim agent policy $\pi^{vic}$, expert policy $\pi^{e}_{MoE}$ |
| **Output**: The optimal adversarial policy network $\pi^{adv}_{\theta^*}$ |
| Initialize network parameters $\theta$ and $\phi$, experience replay buffer $\mathcal{D}_{adv}$ |
| 1:    **For** $iteration = 1, 2, ...$ **do**: |
| 2:       Initialize $\mathcal{D}_{adv}$, $n_1 = \Gamma$ |
| 3:       **While** $\mathcal{D}_{adv}$ not full **do**: |
| 4:          Observe current state $s_t$ and get $a_t = \arg\max_{a \in \mathcal{A}} \pi^{vic}(a \mid s_t)$ |
| 5:          $(x^{adv}_t, a^{adv}_t) \sim \pi^{adv}(s_t, a_t, n_t)$ |
| 5:          **If** $x^{adv}_t > 0$ **and** $n_t > 0$ **then** |
| 6:             $n_{t+1} = n_t - 1$ |
| 7:             $\delta = PerturbationGenerate(s_t, a_t, a^{adv}_t)$ |
| 8:             $s'_t = s_t + \delta$ |
| 9:          **Else** |
| 10:            $s'_t = s_t$, $n_{t+1} = n_t$ |
| 11:          $a^{act}_t = \pi^{vic}(s'_t)$ |
| 12:          Perform $a^{act}_t$ in $Env$ and obtain $(s_{t+1}, \overline{r}_t, done)$ |
| 13:          Store $\tau_t = (s_t, (x^{adv}_t, a^{adv}_t), s_{t+1}, \overline{r}_t, done)$ into $D$ |
| 14:          **If** $n_{t+1} = 0$ **or** $done$ **then** |
| 15:             Reset the environment $\mathcal{E}$ |
| 16:       **End while** |
| 17:       Compute advantage estimates $\hat{A}_t$ |
| 18:       Update $\beta_e$ according to (20) |
| 19:       **For** $epoch = 1, 2, ...$ **do**: |
| 20:          Shuffle collected batch and divide into mini-batches |
| 21:          Compute total loss $\mathcal{L}_{total}(\theta, \phi) = \mathcal{L}'_a(\theta) - c_1 \mathcal{L}_c(\phi) + c_2 \mathcal{H}\left[\pi^{adv}_\theta(\cdot \mid s^{adv}_t)\right]$ |
| 22:          Apply gradient-based updates to $\theta$ and $\phi$ |
| 23:       **End for** |
| 24:    **End for** |

## V. EXPERIMENT EVALUATION

To evaluate the effectiveness of the proposed attack framework, we conduct comprehensive experiments on the SUMO simulation platform. The experimental setup, including environment configuration, Markov decision process (MDP) formulation, baseline selection, training procedures, and evaluation metrics, is detailed in Sections V.A–E. Building upon the experimental setup, a set of experiments are conducted to verify the performance of our method, as detailed in Section V.F.

### A. Environment Setting

We evaluate the performance of our method in two critical autonomous simulated on the SUMO platform, as shown in Fig. 2. Fig. 2(a) illustrates an unprotected left-turn scenario (Env-1), where the DRL-based autonomous driving policy guides the ego vehicle (the red car) to navigate a left turn at an intersection without traffic lights. Fig. 2(b) depicts an on-ramp merging scenario (Env-2), in which the ego vehicle attempts to merge from an on-ramp onto main road. Our method aims to select the optimal timing during the ego vehicle's driving to launch attacks by adding adversarial perturbations that interfere with its actions, causing collisions with other vehicles.

All vehicles in the simulation adopt the LC2013 lane-changing model provided by SUMO. To ensure realistic vehicle dynamics, the vehicle speed is constrained within 15 m/s, and the acceleration is limited to the range of –7.6 m/s² to 7.6 m/s² [20]. The traffic flow density is controlled by the per-second vehicle appearance probability $p$. Unless otherwise specified, this probability is set to $p = 0.5$.

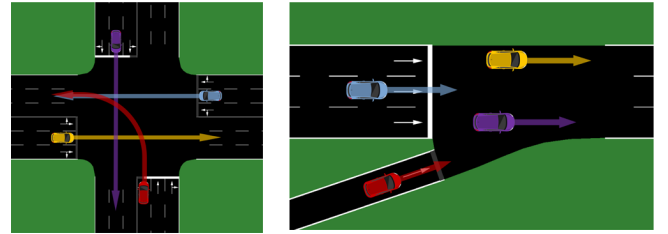

(a) Unprotected left-turn       (b) On-ramp merging
**Fig. 2.** Experimental scenarios.

### B. Markov Decision Process

Based on the simulation environment, we construct an MDP with the following details:

**State Space** $\mathcal{S}$: We assume the state $s_t \in \mathcal{S}$ comprises two main components. The first encodes the ego vehicle's kinematic attributes, including its speed and heading direction. The second component captures the surrounding traffic context by incorporating information from the six nearest vehicles within a 200-meter radius. These vehicles are positioned in six distinct regions relative to the ego vehicle: front, rear, front left, rear left, front right, and rear right. For each adjacent vehicle, the state information comprises its relative distance, orientation angle, speed, and velocity direction. The adversary's state additionally includes two extra dimensions: the remaining number of attacks and the agent's original (pre-perturbation) action.

**Action Space** $\mathcal{A}$: The action space of the autonomous driving agent is defined as continuous acceleration or deceleration within the predefined acceleration range. The adversary's action also includes this component, representing the desired action it intends the agent to take. In addition, an extra binary dimension is introduced to control whether to launch an attack at the current time step.

**Reward function** $r$: The reward function of the autonomous driving agent $r_t$ consists of two components: efficiency reward and safety penalty, which can be expressed as:

$$r_t = \frac{v_t}{v_{max}} - c(s_t, \tilde{a}_t),$$
$$c(s_t, \tilde{a}_t) = \begin{cases} 1, & \text{if collision,} \\ 0, & \text{else,} \end{cases} \quad (23)$$

where $v_{max}$ denotes the maximum speed and $\tilde{a}_t$ represents the agent's actual action at time step $t$, based on the perturbed state if an attack is applied. Specifically, the agent seeks to maximize driving efficiency without compromising safety. To model a worst-case oriented adversary, we design its reward function to solely encourage collisions, thereby driving the agent toward unsafe and failure-prone behaviors. Specifically, we use a binary reward function based on the occurrence of collisions caused by the adversary, where $\bar{r} = c'(s_t, \tilde{a}_t)$. There exists a difference between $c'(\cdot)$ and $c(\cdot)$. Specifically, the adversary receives a reward only when an attack is launched that results in a collision. This design ensures that collisions occur exclusively due to adversarial actions rather than the agent's inherent behavior. Moreover, this concise design also improves cross-scenario adaptability of our method, as the reward depends on a single factor and remains unaffected by other variables.

*C. Baselines*

To validate the effectiveness of our method, we selected the following baselines, including vanilla RL approaches as well as other expert-guided RL methods:

**Vanilla RL (Vanilla)** [48]: The method trains a DRL-based adversary without expert guidance.

**Value Penalty RL (VPRL)**: Building on [17], this method incorporates expert knowledge as a value penalty. Specifically, the method introduces the KL divergence between the expert and adversary policies as a regularization term in the reward function to guide training.

**Policy Constrained RL (PCRL)**: Also based on [17], this method formulates policy optimization as a constrained problem with a KL divergence constraint. The problem is then solved via its Lagrangian dual formulation.

For a fair comparison, all above methods are implemented using PPO as the underlying algorithm. Nevertheless, our method remains agnostic to the choice of DRL algorithm. PPO is selected primarily for its popularity and empirical success in recent reinforcement learning studies.

The autonomous driving agents are based on state-of-the-art reinforcement learning algorithms, including the on-policy method PPO, as well as the off-policy methods SAC [49] and TD3 [50].

*D. Training Details*

The training details are organized into three components: the autonomous driving agents, the MoE-based expert, and the adversarial attack methods.

**Autonomous driving agents**: All algorithms are based on Stable Baselines3 [52]. Each victim agent is trained for 14,000 timesteps using the default hyperparameters provided by Stable Baselines3, with the maximum episode length set to 30.

**MoE-based expert**: The MoE-based expert is an ensemble of five MoE networks, each independently trained using a different random seed. Each MoE network comprises three experts. Training is performed on an expert demonstration dataset containing 250,000 samples for up to 500 epochs, using a batch size of 128 and a learning rate of 0.0003. Early stopping was applied based on validation performance, with training terminated if the mean squared error failed to improve over 20 consecutive epochs.

**Adversarial attack methods**: Each attack method is trained for 512,000 timesteps, using the default PPO hyperparameters provided by Stable Baselines3. For VPRL, the KL divergence coefficient is set to 0.1. For PCRL, the KL divergence tolerance is set to 0.5, and the Lagrangian multiplier is initialized as 0.01. For perturbation generation methodology, we adopt the widely used Basic Iterative Method (BIM) [53][54], which generates adversarial examples through iterative gradient-based updates. The perturbation budget ranges from 0.05 to 0.1, with the number of iterations fixed at 50. The step size is determined by dividing the perturbation budget by the total number of iterations. All methods are evaluated using five independently trained models, each initialized with a different random seed, to ensure robustness and statistical reliability.

All the experiments are conducted on a server equipped with Intel(R) Xeon(R) Gold 6230 CPUs and NVIDIA GeForce RTX 4090 GPUs.

*E. Metrics*

To evaluate the performance of our attack method, we adopt a comprehensive set of metrics commonly used in prior studies. These include success rate (SR), collision rate (CR), average speed (AS), average reward (AR), average number of attacks (ANA), and attack efficiency (AE) [13]. AE is defined as a weighted combination of CR and ANA, designed to quantify the payoff of single-step adversarial attacks. The metric is computed as follows:

$$AE = CR \cdot e^{-\omega \cdot ANA}, \quad (24)$$

where $\omega$ is a custom coefficient, with its value set to 0.05 in our experiments.

*F. Performance Evaluation*

In this section, we present a comprehensive evaluation of our proposed expert-guided adversarial reinforcement learning framework. Unless otherwise stated, all results are derived from models independently trained with five different random seeds.

We first evaluate the performance of different attack methods under the setting of an imperfect expert. We first evaluate the expert's performance under different perturbation budgets and attack budgets, as shown in Table I. In this evaluation, the victim agent is PPO-based, we observe consistent patterns when applying the same expert policy against other DRL-based agents. In Env-1, both the collision rate (CR) and average number of attacks (ANA) increase as $\epsilon$ and $\Gamma$ increase. However, the expert performs less effectively overall, especially under lower attack budgets.

TABLE I.

CR, AND ANA OF THE EXPERT UNDER DIFFERENT CASES

|  | Env-1 | | Env-2 | |
| --- | --- | --- | --- | --- |
|  | CR | ANA | CR | ANA |
| $\epsilon = 0.05, \Gamma = 4$ | 0.51 | 3.53 | 0.16 | 3.93 |
| $\epsilon = 0.05, \Gamma = 7$ | 0.67 | 4.82 | 0.53 | 6.14 |
| $\epsilon = 0.1, \Gamma = 4$ | 0.69 | 3.45 | 0.08 | 3.94 |
| $\epsilon = 0.1, \Gamma = 7$ | 0.83 | 4.35 | 0.46 | 6.36 |

The expert policy is notably suboptimal when $\epsilon = 0.05$ and $\Gamma = 4$. Accordingly, we adopt this setting to evaluate the performance of our method in the presence of imperfect expert guidance. As summarized in Table II, the results demonstrate that our method consistently outperforms existing baselines in terms of CR and AE across most cases. Although the expert policy is imperfect, it can still provide useful guidance during the early stages of training. As a result, both VPRL and PCRL outperform the vanilla attack method, achieving average improvements of 7.24% and 6.56% in collision rate. As our method takes a step further by incorporating the performance-aware annealing strategy, our method achieves over 10% improvement in both collision rate and attack efficiency compared to the vanilla baseline.

Having established the efficacy of our method in low-budget scenarios where the expert policy is imperfect, we now investigate the robustness of our method across multiple scenarios. Specifically, we evaluate our method across diverse environments and against different victim agents, while systematically varying the attack and perturbation budgets. In these new settings, higher budgets typically allow the expert policy to perform better, thereby providing stronger guidance to the adversary. As illustrated in Fig. 3, our method continues to outperform the baselines across a wide range of settings, maintaining high collision rates and low variance. This demonstrates that our method remains robust and effective even under relaxed constraints, where the expert policy provides more reliable guidance, further highlighting the adaptability of our method.

TABLE II.

STATISTICAL RESULTS OF ATTACK METHODS UNDER DIFFERENT SCENARIOS

| Method | Metrics | PPO | | SAC | | TD3 | |
| --- | --- | --- | --- | --- | --- | --- | --- |
|  |  | Env-1 | Env-2 | Env-1 | Env-2 | Env-1 | Env-2 |
| No-attack | CR | 0.02 | 0.00 | 0.01 | 0.01 | 0.01 | 0.00 |
|  | ANA | - | - | - | - | - | - |
|  | AE | - | - | - | - | - | - |
| Vanilla | CR | 0.547 ± 0.097 | 0.825 ± 0.006 | 0.723 ± 0.129 | 0.698 ± 0.040 | 0.392 ± 0.229 | **0.655 ± 0.033** |
|  | ANA | 2.875 ± 0.395 | 2.787 ± 0.043 | 2.822 ± 0.098 | 2.782 ± 0.352 | 3.310 ± 0.142 | 2.507 ± 0.731 |
|  | AE | 0.475 ± 0.086 | 0.718 ± 0.006 | 0.628 ± 0.113 | **0.607 ± 0.026** | 0.334 ± 0.197 | **0.578 ± 0.037** |
| VPRL | CR | 0.535 ± 0.070 | 0.762 ± 0.052 | 0.785 ± 0.010 | 0.682 ± 0.029 | 0.798 ± 0.005 | 0.597 ± 0.028 |
|  | ANA | 3.258 ± 0.235 | 2.585 ± 0.221 | 2.670 ± 0.154 | 2.625 ± 0.196 | 2.777 ± 0.216 | 1.590 ± 0.191 |
|  | AE | 0.455 ± 0.064 | 0.670 ± 0.038 | 0.687 ± 0.013 | 0.598 ± 0.019 | 0.694 ± 0.008 | 0.552 ± 0.023 |
| PCRL | CR | 0.615 ± 0.115 | 0.703 ± 0.022 | 0.768 ± 0.039 | 0.698 ± 0.019 | 0.782 ± 0.029 | 0.542 ± 0.053 |
|  | ANA | 3.167 ± 0.385 | 2.357 ± 0.146 | 2.695 ± 0.154 | 2.828 ± 0.168 | 3.038 ± 0.036 | 3.520 ± 0.107 |
|  | AE | 0.526 ± 0.107 | 0.624 ± 0.019 | 0.671 ± 0.035 | 0.605 ± 0.013 | 0.672 ± 0.025 | 0.455 ± 0.047 |
| Ours | CR | **0.637 ± 0.103** | **0.825 ± 0.031** | **0.812 ± 0.039** | **0.703 ± 0.025** | **0.805 ± 0.006** | 0.560 ± 0.059 |
|  | ANA | 3.138 ± 0.391 | 2.720 ± 0.044 | 2.655 ± 0.166 | 2.992 ± 0.168 | 2.933 ± 0.238 | 3.067 ± 0.911 |
|  | AE | **0.546 ± 0.098** | **0.720 ± 0.028** | **0.712 ± 0.040** | 0.605 ± 0.017 | **0.695 ± 0.013** | 0.482 ± 0.065 |

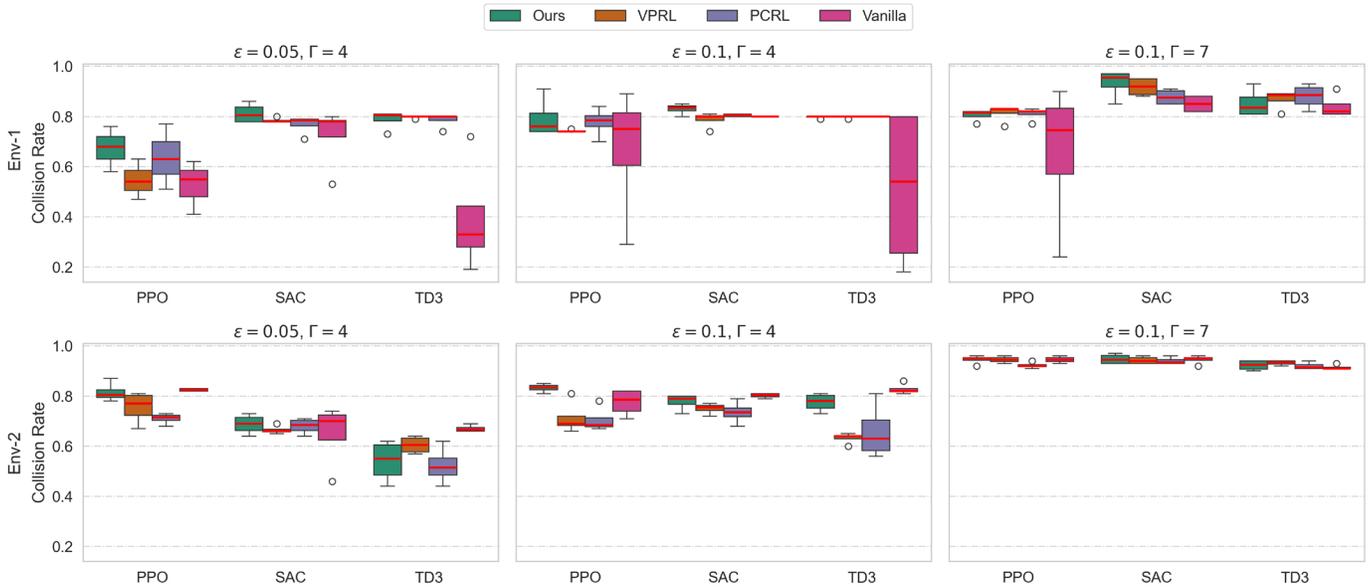

**Fig. 3.** Evaluation of Attack Methods Across Different Attack Budgets and Perturbation Budgets.

## VI. CONCLUSION AND FUTURE WORK

In this work, we presented an expert-guided adversarial attack method against autonomous driving agents. Our method enhances the training stability and effectiveness of DRL-based adversaries under strict attack budget constraints. First, we leverage imitation learning to derive an expert policy from successful attack demonstrations, enabling the policy to capture key decision patterns and behaviors from expert data for more effective guidance. We further incorporate an ensemble Mixture of Experts framework to enhance the adaptability and generalization of the expert policy across diverse scenarios. Then, during the expert-guided training stage, we introduce the KL divergence between the adversary's policy and the expert policy as a constraint to guide the adversary's training process. Furthermore, we propose a novel performance-aware annealing strategy to mitigate reliance on imperfect experts. Experimental results across multiple autonomous driving scenarios show that our method consistently outperforms existing baselines in terms of collision rate, attack efficiency, and convergence stability. Ablation studies show that both our performance-aware annealing strategy and ensemble MoE architecture are highly effective. Specifically, the annealing strategy prevents over-reliance on the expert and promotes more autonomous policy optimization, while the MoE framework enhances adaptability and generalization across diverse scenarios. The effectiveness and stability of our method enable targeted improvements and rigorous evaluation, thereby advancing the safety and robustness of DRL-based autonomous driving policies.